\documentclass[a4paper, 10pt, conference]{ieeeconf}      
\usepackage{FG2021}

\FGfinalcopy 

\IEEEoverridecommandlockouts                              
\usepackage{graphics} 
\usepackage{epsfig} 
\usepackage{mathptmx} 
\usepackage{times} 
\usepackage{amsmath} 
\usepackage{amssymb}  
\usepackage{hyperref}
\usepackage{subfigure}
 \usepackage{array,multirow,graphicx}
\usepackage[accsupp]{axessibility} 

\def\FGPaperID{****} 

\title{\LARGE \bf
Mask-invariant Face Recognition through Template-level Knowledge Distillation
}

\author{\parbox{16cm}{\centering
    {\large Marco Huber$^1$$^,$$^2$, Fadi Boutros$^1$$^,$$^2$, Florian Kirchbuchner$^1$, Naser Damer$^1$$^,$$^2$}\\
    {\normalsize
    $^1$ Fraunhofer Institute for Computer Graphics Research IGD, Darmstadt, Germany\\
    $^2$ Department of Computer Science, TU Darmstadt, Darmstadt, Germany}}
    \thanks{This research work has been funded by the German Federal Ministry of Education and Research and the Hessian Ministry of Higher Education, Research, Science and the Arts within their joint support of the National Research Center for Applied Cybersecurity ATHENE.}
}

\begin{document}

%
%
%




\IEEEoverridecommandlockouts\pubid{\makebox[\columnwidth]{978-1-6654-3176-7/21/\$31.00~\copyright{}2021 IEEE \hfill}
\hspace{\columnsep}\makebox[\columnwidth]{ }}

\ifFGfinal
\thispagestyle{empty}
\pagestyle{empty}
\else
\author{Anonymous FG2021 submission\\ Paper ID \FGPaperID \\}
\pagestyle{plain}
\fi
\maketitle

\begin{abstract}
The emergence of the global COVID-19 pandemic poses new challenges for biometrics. Not only are contactless biometric identification options becoming more important, but face recognition has also recently been confronted with the frequent wearing of masks. 
These masks affect the performance of previous face recognition systems, as they hide important identity information. 
In this paper, we propose a mask-invariant face recognition solution (MaskInv) that utilizes template-level knowledge distillation within a training paradigm that aims at producing embeddings of masked faces that are similar to those of non-masked faces of the same identities. 
In addition to the distilled knowledge, the student network benefits from additional guidance by margin-based identity classification loss, ElasticFace, using masked and non-masked faces.
In a step-wise ablation study on two real masked face databases and five mainstream databases with synthetic masks, we prove the rationalization of our MaskInv approach. 
Our proposed solution outperforms previous state-of-the-art (SOTA) academic solutions in the recent MFRC-21 challenge in both scenarios, masked vs masked and masked vs non-masked, and also outperforms the previous solution on the MFR2 dataset. 
Furthermore, we demonstrate that the proposed model can still perform well on unmasked faces with only a minor loss in verification performance.
The code, the trained models, as well as the evaluation protocol on the synthetically masked data are publicly available: \url{https://github.com/fdbtrs/Masked-Face-Recognition-KD}.
\end{abstract}

\section{Introduction}
The current COVID-19 pandemic presented new challenges to biometric technologies \cite{martaERA}. To reduce the risk of spreading the virus, the use of contactless biometrics has become increasingly important. Face recognition (FR) in particular had already established itself as a contactless biometric modality before the pandemic due to its performance \cite{BoutrosMFR21} and its passive, universal and non-intrusive nature \cite{JainRP04}. However, FR systems have also been confronted with new realities, most notably the wearing of masks by the general public to prevent the spread of the contagious virus. The presence of a mask that hides facial features can weaken the FR system \cite{FRVT, damer2021extended, DamerGCBKK20} and thus reduce confidence in the system's decisions \cite{martaERA}. A study by the National Institute of Standards and Technology (NIST) \cite{FRVT} as well as a study from the Department of Homeland Security \cite{batagelj} concluded, that the wearing of masks has a significant negative effect on the accuracy of FR systems. Further studies from the scientific community confirmed this \cite{damer2021extended, DamerGCBKK20}. It is important to note that the negative effect of masks affects not only the recognition performance of automated FR systems but also the performance of human operators \cite{humans}, presentation attack detection \cite{FANG2021108398}, and quality estimation \cite{FuMaskedQuality}. Therefore, it is important to find technical solutions to make FR systems robust to  wearing masks in response to the given circumstances. The importance of this topic has recently led to the organization of several competitions addressing masked face recognition \cite{zhu2021masked, deng2021masked, BoutrosMFR21}.

The challenge of masked faces for automatic FR  came into scientific focus with the onset of the COVID-2019 pandemic. Previously, some work generally dealt with the problem of occluded faces, which also included sunglasses, other facial accessories, and other face occlusions \cite{JiaM09, occ18, vitomir}. Also in the same direction, the detection (not the recognition) of occluded faces or masked faces \cite{Ge2017CVPR, batagelj,loey} was also discussed in the literature. Such works addressed the problem of detecting faces that are largely occluded or faces that are partially covered by masks. There were only a few works that address enhancing the recognition performance on masked faces. Li et al. \cite{li_attention} proposed a cropping and attention-based approach to train a face recognition model on the periocular area of masked faces. Neto et al. \cite{pedro} also focus on the upper part of the face, for this, they use a constraint triplet loss to get optimized embeddings for masked face recognition. Another approach that has been followed is to synthetically replace the area hidden by the mask using Generative Adversarial Networks (GANs) \cite{gan1, LiMasks, gan2}. Li et al. \cite{LiMasks} proposed a solution to in-paint the masked area while trying to maintain the identity information. A different approach was taken by Boutros et al. \cite{triplet}, where the authors proposed a self-restrained triplet loss to train an on-the-top solution that learns to transfer the templates of masked faces into templates that possess the properties of non-masked face templates. Several approaches were proposed as part of the MFRC-21 Challenge \cite{BoutrosMFR21}. Most of the submissions used a ResNet \cite{resnet100} architecture and ArcFace \cite{arcface} loss as a foundation. Anwar and Raychowdhury \cite{masktheface} trained a model on synthetically masked images using the Inception-ResNet v1 \cite{inception} with the triplet-loss FaceNet \cite{facenet}. Moreover, Zheng et al. \cite{web} used large-scale web-collected database and corresponding tags without manual annotations and used frequency domain information to train an FR that is more robust to masked faces.

\begin{figure*}[th]
    \centering
    \includegraphics[width=0.70\textwidth]{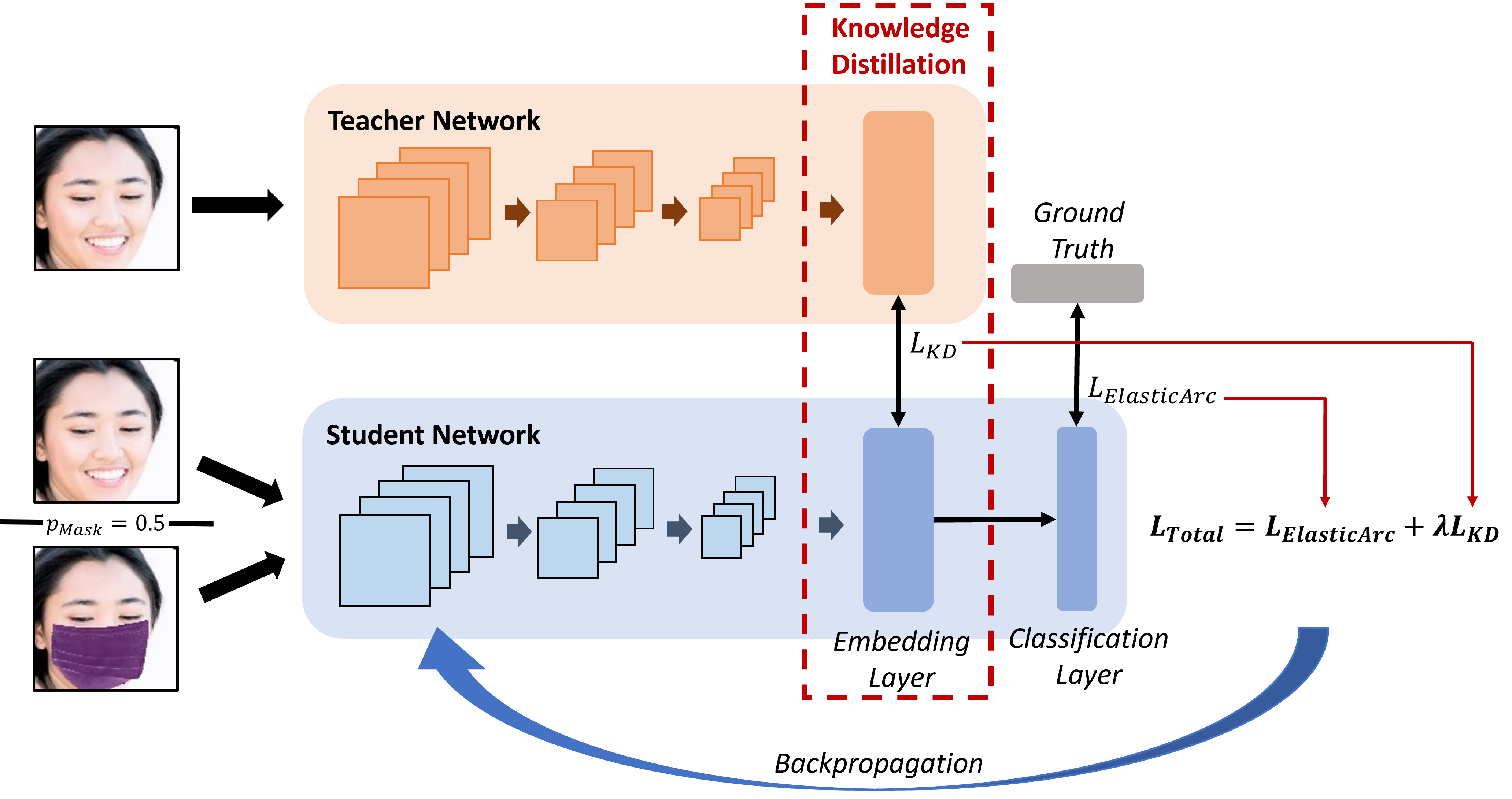}
     \caption{\textbf{Overview of the proposed MaskInv approach for Masked Face Recognition:} The pre-trained teacher network and the student network to be trained are forwarded the same images. With a probability $p_{Mask}$ of 0.5, a synthetic mask is added to the input image of the student network. During training, the mean squared error ($L_{KD}$) is calculated between the face embedding of the teacher and the student network. The calculated error then contributes to the overall loss $L_{Total}$ of the student network that is trained additionally with ElasticFace-Arc loss ($L_{ElasticArc}$) \cite{elasticface}.}
    \label{overview}
    \vspace{-3mm}
\end{figure*}

Knowledge Distillation (KD) is a technique that is commonly used to improve the performance and generalizability of lightweight models. This is achieved by transferring knowledge learned by a teacher model to a (usually) smaller student model. The student model is guided by the teacher model to learn additional relationships discovered by the teacher model that goes beyond the information stored in ground truth labels \cite{distilling}. For Face Recognition, KD has already been used to reduce the complexity of face recognition models to produce well-performing lightweight models \cite{pocketnet, margin}, or to counter the problem of low-resolution FR \cite{Wang_2019_ICCV, kdlow}. Li et al. \cite{LiMasks} used KD between a teacher and a student model to distill the feature distribution of unmasked faces to recovery identity information on inpainted masked faces.

In this work, we propose a mask-invariant face recognition solution, namely the MaskInv. MaskInv utilizes knowledge from a pre-trained face recognition model using KD on template-level while also being guided by a margin-based identity classification loss (ElasticFace \cite{elasticface}).
The student model is trained with masked and non-masked face images so that it can deal with both cases, while the KD process ensures that the model would produce embeddings of masked faces that are similar to those of non-masked faces of the same identities.
We investigated single and two-stage training paradigm, where the latter puts more emphasis on the KD at later training stages proving to enhance the masked FR performance.
We additionally baseline our solution, to the same training paradigm that includes using face image augmented with masks, however, without optimizing the embeddings using KD.

Our experiments demonstrate, in a stepwise ablation study, the accuracy gains of our MaskInv solution on different scenarios of masked face recognition. 
This study utilizes seven different benchmarks, two with real and five with simulated masked face images. We further prove the superiority of our MaskInv solution by comparing the achieved results to the top academic performers of the MFR 2021 masked face recognition challenge \cite{BoutrosMFR21}, where the MaskInv outperformed all the academic solutions in both, masked vs masked and masked vs non-masked scenarios. 

The paper is structured as follows: in Section \ref{meth} we detail and rationalize our proposed approach. The experimental setup, the databases used for training and evaluation, and the evaluation criteria are described in Section \ref{exp}. Subsequently, we present and discuss our results in Section \ref{results}, both in terms of a detailed ablation study and comparison with SOTA. In Section \ref{concl} we conclude our work.

\vspace{-1mm}
\section{Methodology}
\label{meth}
In this section, we present and rationalize our proposed methodology to create mask-invariant face embeddings through our MaskInv solution. Our approach achieves that by jointly learning the correct identity classification of masked and non-masked face images, and ensuring that the embeddings of masked faces are similar to those of non-masked images of the same identities through embedding-level KD. The KD teaches a student network to process a masked face in a manner that produces an embedding similar to the non-masked face embedding produced by the teacher, and thus try to neglect the non-identity related information introduced by the mask.

A well-performing face recognition model trained on non-masked faces acts as the teacher model in our knowledge-distillation architecture. A second FR system acts as the student network and is trained with interaction to the teacher model to be mask-invariant, and thus produce embeddings from masked faces that are similar to those produced from non-masked faces. A schematic overview of the proposed learning scenario is presented in Figure \ref{overview}. During the training, the same images are simultaneously fed to both, the teacher and the student model. On the images forwarded to the student model, we apply a synthetic mask with a probability $p_{Mask}$, while the images forwarded to the teacher network remain unaltered. The synthetic mask is created by placing a mask template on the face, depending on the landmarks used for the face alignment during the image pre-processing. The synthetic mask is applied to a proportion of the images fed to the student model (probability $p_{Mask}$) to ensure that it still deals optimally with non-masked faces and to enable a more stable training process. For the teacher network, we use a pre-trained auxiliary network that guides the newly trained student network. To achieve our goal, the student network is not only trained to produce a correct classification decision but additionally optimize the produced embedding to be similar to that of the teacher network (on non-masked faces).
Thus, the student network is trained using a combined loss $L_{Total}$, consisting of two different losses. Formally, we define the total loss $L_{Total}$ as:
\vspace{-1mm}
\begin{equation}
    L_{Total} = L_{ElasticArc} + \lambda L_{KD},
    \label{equ:tloss}
    \vspace{-1mm}
\end{equation}
where $L_{ElasticArc}$ refers to the recently published SOTA FR loss function ElasticFace-Arc \cite{elasticface} and $L_{KD}$ to the mean squared error loss in the template-level KD process. The ElasticFace-Arc loss function relaxes the fixed margin constraint of similar high-performing FR loss functions and therefore provides space for flexible identification separability. It outperforms several other SOTA FR loss functions especially at hard cross-pose benchmarks \cite{elasticface}. Formally, it can be defined as \cite{elasticface}:
\begin{equation}
\resizebox{\linewidth}{!}{$
    L_{ElasticArc}= \frac{1}{N}  \sum\limits_{i \in N} - log \frac{e^{s (cos(\theta_{y_i}+E(m,\sigma)))}}{ e^{s (cos(\theta_{y_i}+E(m,\sigma)))} +\sum\limits_{j=1 , j \ne y_i}^{c}  e^{s ( cos(\theta_{y_j}))}},$}   
\end{equation}
where $N$ denotes the batch size, $c$ the number of identities, $s$ the scale parameter, $m$ the margin parameter, and $\sigma$ the standard deviation. The function $E(m,\sigma)$ returns a random value from a Gaussian distribution with the mean $m$ and the standard deviation $\sigma$. All the parameters are set as defined in \cite{elasticface}.
The $L_{KD}$ is calculated as part of the KD between the feature embeddings of the teacher network and the feature embeddings of the student network to optimize the embedding itself rather than the network classification behavior. This ensures that the embedding distortions caused by masks are kept to a minimum and therefore guides the KD process to produce a mask-invariant student network. The used $L_{KD}$ loss, mean squared error, can be formalized as: 
\vspace{-2mm}
\begin{equation}
    L_{KD}= \frac{1}{N}  \sum\limits_{i \in N} \Big( 1- \frac{1}{n}\Sigma_{j=1}^{D}{\Big(\Phi^S_i(x)_j -\Phi^T_i(x)_j\Big)^2\Big)},
    \vspace{-1mm}
\end{equation}
where $\Phi^S_i$ and $\Phi^T_i$ are the feature representations obtained from the embedding layer of the student and teacher model, respectively, and $D$ is the size of the embeddings.

Since the learned feature embeddings are normalized, the range of $L_{KD}$ is rather small and thus, as proposed by \cite{pocketnet}, we weight them with a weight $\lambda = 100$ during the first training step. This enables the knowledge transfer by allowing the $L_{KD}$ to contribute to the overall loss while keeping the emphasis on learning identity classification by the $L_{ElasticArc}$.

We propose two different paradigms based on the methodology described above. In the MaskInv-HG (Mask-invariant High Guidance) approach, we increased the weighting to $\lambda = 3000$ when the $L_{ElasticArc}$ loss stabilized to further emphasize the adaption of the network to the masked data. In the MaskInv-LG (Mask-invariant Low Guidance) approach the weighting remains unchanged throughout the training process. 
For a detailed ablation study, we additionally present the results of the third solution, where the $\lambda$ is set to zero, and thus no KD is applied, rather the student network is trained independently with faces augmented with a synthetic mask with the probability $p_{Mask}$, this solution will be referred to as ElasticFace-Arc-Aug. All the training parameters will be introduced in the next section.  

\vspace{-1mm}
\section{Experimental Setup}
\label{exp}
\subsection{Training Setup}
\label{trainingsetup}
As the teacher network, we use a pre-trained FR model based on the ResNet-100 \cite{resnet100} architecture, trained on the MS1MV2 dataset \cite{arcface} with ElasticFace-Arc loss \cite{elasticface}, which has been made publicly available by the authors\footnote{https://github.com/fdbtrs/ElasticFace}. For the training details of the teacher model, we refer to the original ElasticFace paper \cite{elasticface}. We chose this model because it advanced the SOTA (such as ArcFace \cite{arcface} and MagFace \cite{magface}) on six difficult mainstream benchmarks and is publicly available.

For the student model, we also use the ResNet-100 \cite{resnet100} architecture as the ResNet-100 architecture is widely used in SOTA FR approaches \cite{magface,arcface,elasticface}. For the ElasticFace-Arc loss we set the scale parameter $s$  to $s=64$, similar to \cite{arcface, elasticface, curricularface} and the margin parameter $m$ to $m=0.5$ and the standard deviation $\sigma$ to $\sigma=0.5$, following \cite{elasticface}. The mini-batch size is set to 512. The model is trained with Stochastic Gradient Descent (SGD) optimizer with an initial learning rate of $0.1$. The momentum is set to $0.9$ and the weight decay to $5e^{-4}$. The learning rate is divided by 10 at 80k, 140k, 210k training iterations.  The total amount of training iterations is set to 295k iterations. With these parameters and training process, we follow the training process defined in \cite{elasticface}. For the MaskInv-HG variant, we increase the weighting $\lambda$ from 100 to 3000 after 227k iterations. We do this to further sensitize the model to the masked data. For the MaskInv-LG variant, the weighting remains unchanged throughout the training. In the experiments, we investigate the performance of both, the student model with low guidance (MaskInv-LG) and the student model with high guidance (MaskInv-HG), in a detailed ablation study.
For a detailed ablation study, we also evaluate our ElasticFace-Arc-Aug solution where the $\lambda$ is set to zero (no KD). Here, the student network is trained independently (with the same training procedure as the teacher)  with faces augmented with a synthetic mask with the probability $p_{Mask}$.

Following recent trends \cite{arcface, elasticface, magface, curricularface}, the model is trained on the MS1MV2 dataset \cite{arcface}, which is the same data used to train the teacher model. The MS1MV2 is a refined version of MS-Celeb-1M \cite{ms1m} and contains 5.8M images of 85k identities. For the teacher network, the images are used unmodified, while for the student network, with a probability of 0.5, synthetic masks with random colors and random small deviations in shape are added. The synthetic masked images were created by mapping a template mask image on the extracted landmarks used for pre-processing with small variations in the mapped key points. All the images are aligned and cropped to 112x112x3 using MTCNN \cite{mtcnn} and then normalized to have pixel values between -1 and 1. The simulated mask approach will be publicly provided to ensure comparability and reproducibility.

\begin{figure}
    \centering
    \subfigure[]{\includegraphics[width=0.10\textwidth]{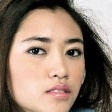}} 
    \subfigure[]{\includegraphics[width=0.10\textwidth]{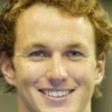}} 
    \subfigure[]{\includegraphics[width=0.10\textwidth]{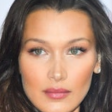}} 
    \subfigure[]{\includegraphics[width=0.10\textwidth]{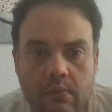}}
    
    \subfigure[]{\includegraphics[width=0.10\textwidth]{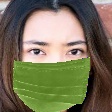}} 
    \subfigure[]{\includegraphics[width=0.10\textwidth]{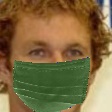}} 
    \subfigure[]{\includegraphics[width=0.10\textwidth]{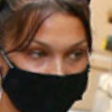}} 
    \subfigure[]{\includegraphics[width=0.10\textwidth]{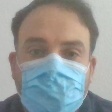}}
    
    \caption{(a, e) Images from the MS1MV2 dataset used for training (b, f) Images from the LFW benchmark (c, g) Images from the MFR2 dataset, (d,h) Images from the MFRC-21 dataset. The masks in (e) and (f) were synthetically added for the training or evaluation purpose on masked images. The images in each column are from the same identity.}
    \label{samples}
    \vspace{-6mm}
\end{figure}

\subsection{Evaluation Dataset \& Benchmarks}
To demonstrate the effect of our proposed masked face recognition approach, we investigate the behavior on both, real and simulated masked face datasets. Furthermore, we observe the performance on non-masked benchmarks to show that our MaskInv solution can still operate well on non-masked data. 

Regarding the performance on real masked data, we use the database MFRC-21 of the MFR competition \cite{BoutrosMFR21} as well as MFR2 \cite{masktheface}. The MFRC-21 dataset consists of images of 47 individuals taken in a collaborative but varying scenario. The authors of MFRC-21 provided two different evaluation scenarios: non-masked vs masked and masked vs masked. The evaluation on this database (MFRC-21) is essential as it enables a wide range comparison to SOTA solutions presented in the competition \cite{BoutrosMFR21}. Additionally to the MFRC-21, the MFR2 dataset \cite{masktheface} is also used in this work. The MFR2 consists of masked images collected in the wild of 53 identities with a total of 269 images. The authors provide a list of 848 pairs of images to be compared.

We additionally opted to investigate the performance of our MaskInv solution on large-scale databases. To do that, we rely on simulated masks to adapt established large-scale face recognition benchmarks into evaluating masked face recognition performance. To generate the simulated mask, we follow the same synthetic mask-creating process detailed in \autoref{trainingsetup}. However, we do not use the random shift in the mask position to maintain a realistic appearance, but we retain the random color of the mask. Following the proposed scenarios of the MFRC-21 competition, we evaluate both, masked vs masked and non-masked vs masked comparison pairs, where we either apply the simulated mask on both images or only on one image in the comparison pair. As a basis for our simulated large-scale masked dataset, we use five mainstream datasets that also cover various additional challenges. The five used benchmarks are: LFW \cite{lfw}, CFP-FP \cite{cfp}, AgeDB-30 \cite{agedb}, CALFW \cite{calfw}, and CPLFW \cite{cplfw}.

LFW \cite{lfw} is an unconstrained face verification benchmark and contains 13,233 images of 5749 different identities. We follow the standard protocol and applied the masks on the defined 6000 comparison pairs. The CFP-FP dataset \cite{cfp}  addresses the comparison between frontal and profile faces and the evaluation protocol contains 3500 genuine and 3500 imposter pairs. AgeDB \cite{agedb} is an in-the-wild dataset for age-invariant face verification. We use the most reported and most challenging scenario AgeDB-30 with an age gap of 30 years between the images of the individuals. Both, the CALFW dataset \cite{calfw} and the CPLFW \cite{cplfw} are based on the LFW dataset. While the CALFW dataset focuses on cross-age evaluation, the CPLFW dataset focus on cross-pose evaluation. Both dataset protocols provide 3000 genuine and 3000 imposter comparison pairs. The imposter pairs of CALFW and CPLFW are selected from the same gender and ethnicity to reduce the effect of these attributes on the recognition performance. 

To sum it up, we use seven databases to evaluate our MaskInv solution, two with real masks and five common face recognition benchmarks with simulated masks. Sample images from the used datasets are shown in Figure \ref{samples}.

\subsection{Evaluation Criteria}
For the evaluation, we consider several metrics. For the sake of comparability and reproducibility, we follow the evaluation metrics used and proposed in the utilized benchmarks and datasets. We nevertheless acknowledge the evaluation metrics defined in the ISO/IEC 19795-1  \cite{mansfield2006information} standard.

The verification performance on the MFRC-21 dataset \cite{BoutrosMFR21} is evaluated and reported as the false non-match rate (FNMR) at two operation points. The operation points are denoted as FMR1000 and FMR100 and refer to the points which provide the lowest FNMR for a false match rate (FMR) $<0.1\%$ and $< 1.0\%$, respectively. As the FMR1000 has been proposed as the best practice evaluation operation point for high-security scenarios, e.g. automatic border control systems by the European Border and Coast Guard Agency (Frontex) \cite{frontex2015best}, we rate it higher than FMR100. To get an indication of generalizability, we also report the Fisher Discriminant Ratio \cite{fdr} between the genuine and imposter distributions as a separability measure, similar to \cite{BoutrosMFR21}. 

On the MFR2 dataset \cite{masktheface} we follow the evaluation metrics reported in the original work. This includes the true positive rate (TPR) at a false acceptance rate (FAR) of $0.2\%$ (FAR2000), the achieved accuracy at this operation point (ACC2000) and the maximum accuracy of the network (ACC) \cite{masktheface}. For the benchmarks LFW \cite{lfw}, CFP-FP \cite{cfp}, AgeDB-30 \cite{agedb}, CALFW \cite{calfw}, and CPLFW \cite{cplfw}, we follow the original metric in the respective benchmarks where they all report the verification performance as the verification accuracy in percentage points as defined in the corresponding benchmark references.

\begin{table}[]
\caption{Ablation Study on the MFRC-21 dataset - Mask vs No-Mask: With the KD and the high guidance (MaskInv-HG), the separability and the performance at FMR100 increased.}
\vspace{-3mm}
\label{abl_mfrc21-nomask}
\begin{tabular}{|l|c|c|c|}
\hline
\multicolumn{1}{|c|}{\textbf{Mask vs No-Mask}}     & FMR1000            & FMR100            & FDR               \\ \hline
ArcFace \cite{arcface}                             & 0.07154            & 0.06009           & 8.6899            \\ \hline
MagFace \cite{magface}                             & 0.07641            & 0.05906           & 8.4147            \\ \hline
ElasticFace-Arc (teacher) \cite{elasticface}       & 0.08112            & 0.06004           & 8.2282            \\ \hline
ElasticFace-Arc-Aug (ours)                         & \textbf{0.05792}   & 0.05692           & 10.8983           \\ \hline
MaskInv-LG (ours)                                  & 0.05951            & 0.05696           & 10.6944           \\ \hline
MaskInv-HG (ours)                                  & 0.05849            & \textbf{0.05637}  & \textbf{11.3271}  \\ \hline
\end{tabular}
\end{table}

\begin{table}[]
\caption{Ablation Study on the MFRC-21 dataset - Mask vs Mask: Similar to the Mask vs Non-Mask scenario, the KD, and the high guidance were beneficial to the separability.}
\vspace{-3mm}
\label{abl_mfrc21-mask}
\begin{tabular}{|l|c|c|c|}
\hline
\multicolumn{1}{|c|}{\textbf{Mask vs Mask}}        & FMR1000            & FMR100            & FDR               \\ \hline
ArcFace \cite{arcface}                             & 0.06504            & 0.05925           & 9.6864            \\ \hline
MagFace \cite{magface}                             & 0.06694            & 0.05794           & 9.8259            \\ \hline
ElasticFace-Arc (teacher) \cite{elasticface}       & 0.07109            & 0.05795           & 9.5872            \\ \hline
ElasticFace-Arc-Aug (ours)                         & \textbf{0.05808}   & \textbf{0.05606}  & 10.9477           \\ \hline
MaskInv-LG (ours)                                  & 0.05923            & 0.05681           & 10.9671           \\ \hline
MaskInv-HG (ours)                                  & 0.05886            & 0.05654           & \textbf{11.5337}  \\ \hline
\end{tabular}
\vspace{-3mm}
\end{table}

\begin{table}[t]
\caption{Ablation Study on the MFR2 dataset: The FAR2000 and ACC2000 increased with the KD and the high guidance. }
\vspace{-3mm}
\label{abl-mfr2}
\begin{tabular}{l|c|c|c|}
\cline{2-4}
                                                                        & FAR2000           & ACC2000       & ACC     \\ \hline
\multicolumn{1}{|l|}{ArcFace \cite{arcface}}                            & 85.61\%           & 92.81\%       & 93.51\%  \\ \hline
\multicolumn{1}{|l|}{MagFace \cite{magface}}                            & 88.92\%           & 94.46\%       & 95.17\% \\ \hline
\multicolumn{1}{|l|}{ElasticFace-Arc (teacher) \cite{elasticface}}      & 83.25\%           & 91.63\%       & 95.05\% \\ \hline
\multicolumn{1}{|l|}{ElasticFace-Arc-Aug (ours)}                        & \textbf{92.45\%}  & \textbf{96.11\%}       & \textbf{96.93\%} \\ \hline
\multicolumn{1}{|l|}{MaskInv-LG (ours)}                                 & 91.98\%           & 95.99\%       & 96.46\% \\ \hline
\multicolumn{1}{|l|}{MaskInv-HG (ours)}                                 & 92.21\%           & \textbf{96.11\%}       & 96.34\% \\ \hline
\end{tabular}
\vspace{-4mm}
\end{table}

\vspace{-2mm}

\vspace{-1mm}
\section{Results}
\label{results}
In this section, we present the evaluation results achieved by our MaskInv solution. We start with an extended ablation study to investigate the influence of our proposed solution in its two training paradigms MaskInv-LG and MaskInv-HG in comparison to the teacher baseline (ElasticFace-Arc) and the ElasticFace-Arc-Aug (trained without KD). Later on, we take a closer look at the performance of our approach in comparison to results published in the literature.
\vspace{-1mm}
\subsection{Ablation Study} 
\begin{figure*}[t]
    \centering 
    \subfigure[]{\includegraphics[width=0.40\textwidth]{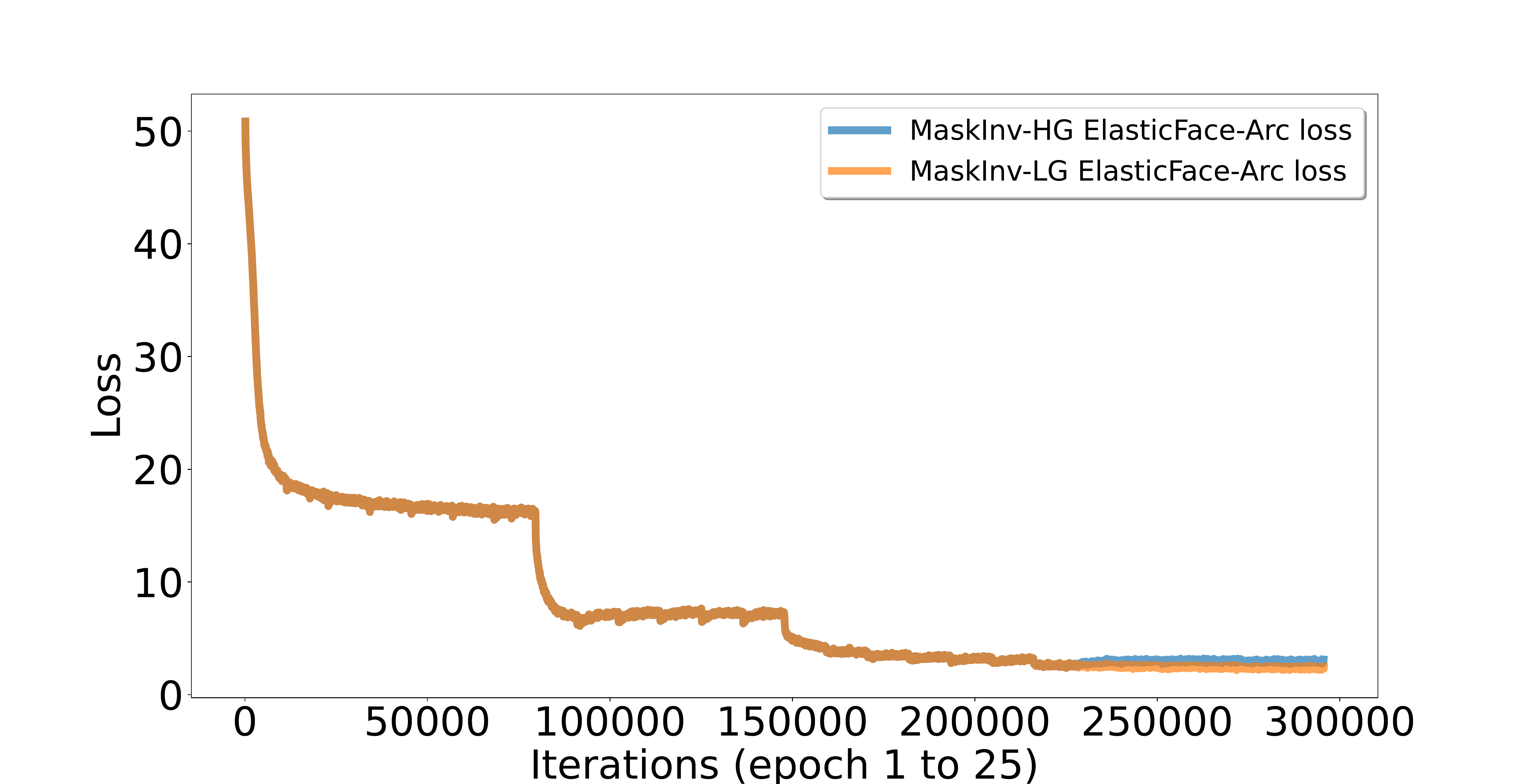}} 
    \subfigure[]{\includegraphics[width=0.40\textwidth]{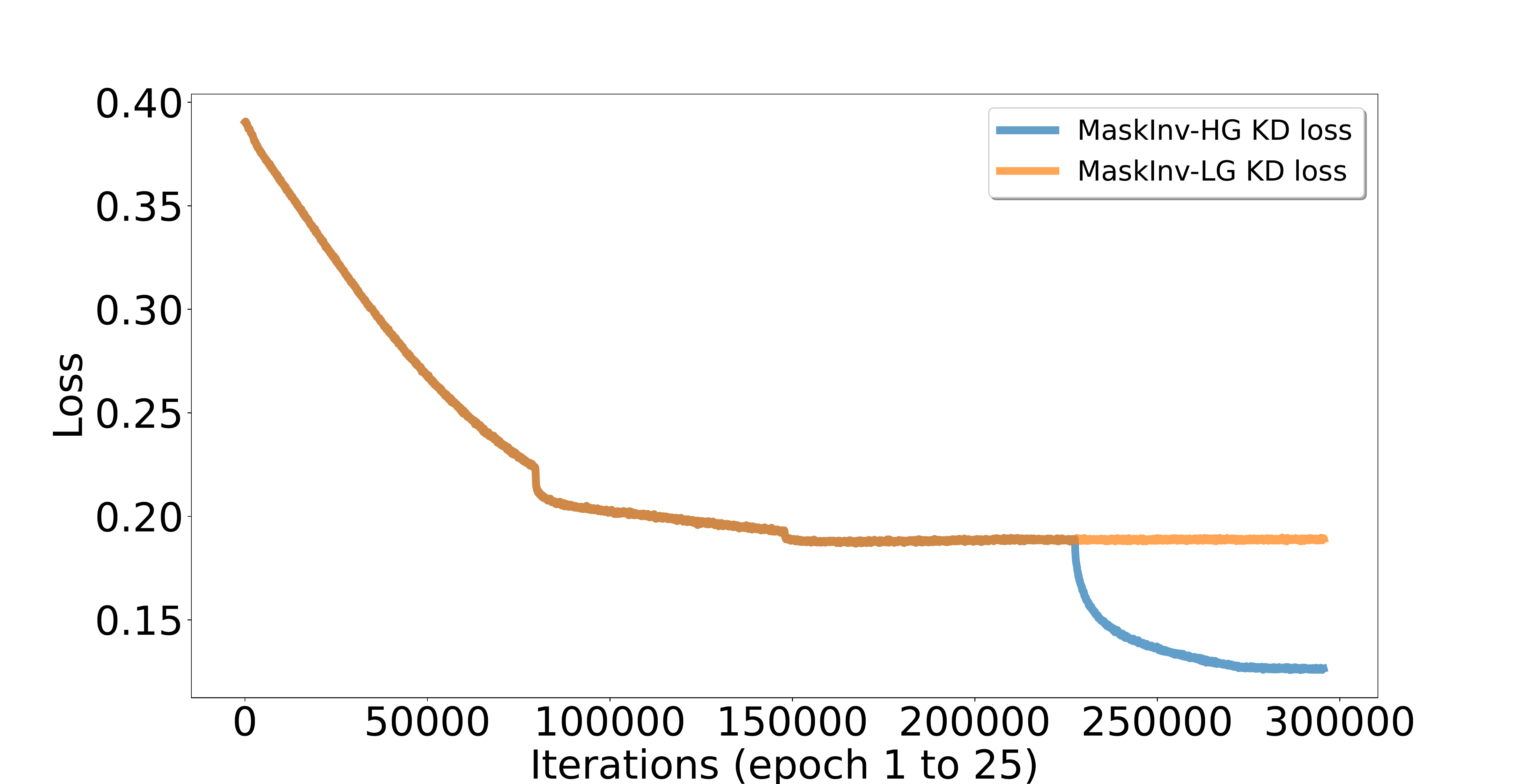}} 
    \vspace{-2mm}
    \caption{The ElasticFace-Arc loss (a) and the KD loss (b) for the MaskInv-HG and MaskInv-LG training paradigms. For the MaskInv-HG, the $\lambda$ (is Equation \ref{equ:tloss} is increased from 100 to 3000 after 227k iterations. This change results in only a minimum change in the the ElasticFace-Arc loss, but a huge drop in the embedding optimization loss (KD loss), when compared to the  MaskInv-LG training paradigm where the $\lambda$ is unchanged.
    }
    \vspace{-6mm}
    \label{losses}
\end{figure*}

In the following ablation study, we investigate step by step the impact of our two-stage training paradigm by looking into the (1) teacher baseline, (2) the model trained with mask-augmented images but no KD ($\lambda=0$), (3) the single step MaskInv-LG, and (4) the 2 stage MaskInv-HG solutions. 
We additionally, present our ablation result study in perspective of two additional (to ElaticFace-Arc) top-performing face recognition solutions (ArcFace \cite{arcface}, MagFace \cite{magface}, both also based on ResNet-100 and trained on MS1MV2 \cite{arcface}) to motivate mask-specific solutions, while we maintain the comparison to SOTA solutions that specifically targeted masked face recognition in the next section.
The results on the MFRC-21 dataset of the different models are shown in \autoref{abl_mfrc21-nomask} and \autoref{abl_mfrc21-mask} for the masked vs non-masked and the masked vs masked scenarios, respectively. 
We focus our discussion on the verification performance FMR1000 as this has been proposed by Frontex as a best practice operation point for processes just as automatic border control \cite{frontex2015best}. 
\autoref{abl_mfrc21-nomask} and \autoref{abl_mfrc21-mask} show that the KD is beneficial to the verification performance in comparison to the teacher model and that the adjusted models outperform the traditional FR models on masked faces. 
While the ElasticFace-Arc-Aug shows in few cases a better performance in comparison to MaskInv-HG, it lacks the same level of separability, measured in FDR, and thus the expected generalizability of the performance. 
\autoref{abl-mfr2} presents a similar ablation study on the MFR2 database, where again the proposed MaskInv-HG presented superior performance to the teacher and the MaskInv-LG model by scoring an FAR2000 of 92.21\%, in comparison to 91.98\%, and 83.25\% for the MaskInv-LG, and the teacher network, respectively. A similar trend can be seen for ACC2000 metric. The maximum accuracy ACC shows mixed results, however such a metric is not measured at a comparable operation point between different solutions and thus can not lead to a fair comparison, which is also the reason why such metrics are not included in the verification performance metrics in ISO/IEC 19795-1  \cite{mansfield2006information}. 
The poorer performance of the two well-known MagFace and Arcface-based models on the real masked datasets (MFRC-21 and MFR2) shows that a tailored solution, such as the one we are presenting here) for MFR is necessary.



In \autoref{abl-bench} we present the results on five face recognition benchmarks. The experiments were performed on three different versions of the datasets: 1) on the unaltered data (No Mask), 2) on the dataset with synthetics masks applied on one of the images of each pair (Masked vs Non- Masked), 3) on the dataset with synthetic masks on both images of each pair (masked vs masked). The results from \autoref{abl-bench} show that all three models are better at handling unmasked data than masked data and outperform conventional FR models in the latter case. 
The proposed MaskInv solutions guided by the KD outperform the teacher model as well as the ElasticFace-Arc-Aug model on most synthetic mask benchmarks. However, the MaskInv-LG and MaskInv-HG do iterate on the top performance spot. For example, on the masked vs non-masked settings of the CFP-FP benchmark, the MaskInv-HG comes first with an accuracy of 96.94\%, followed by 96.83\% and 95.91\% by the MaskInv-LG and ElasticFace-Arc-Aug, respectively. However, all far ahead of the teacher and other baseline methods. Similar scenarios can be seen in the masked vs masked setting, with the ElasticFace-Arc-Aug edging closer to the MaskInv solutions. This is the case as the masked vs masked setting benefits less from the main goal of the MaskInv solution, which is to create face representations similar between masked and unmasked faces, while the masked vs masked setting require only the similarity between masked faces. 
The lower masked face recognition performances of the solutions based on MagFace, ArcFace, and ElasticFace-Arc (teacher) again indicate the need for specifically designed solution as our MaskInv.


To examine the influence of the proposed training paradigm in detail, we present the plot of the two losses in Figure \ref{losses} plotted along the training iterations. 
When comparing to the MaskInv-LG, the $\lambda$ (see Equation \ref{equ:tloss}) is increased for the MaskInv-HG from 100 to 3000 after 227k iterations resulting in a minimum change in the the ElasticFace-Arc loss, but a huge drop in the embedding optimization loss (KD loss). This corresponds to the targeted effect of not effecting the identity classification performance of the model, while enhancing the similarity of the embedding (whether masked or not) to that of the teacher model.


In summary, the ablation study showed that our proposed MaskInv solution benefits from learning to produce similar embeddings for masked and non-masked faces through knowledge distillation. This is specifically beneficial when comparing masked to non-masked faces as demonstrated and rationalized earlier, which is the more practical comparison scenario (e.g. non-masked reference in passport). 
The experiments performed on benchmarks addressing special challenges such as cross-pose and cross-age, show that the proposed approach also can be transferred to these cases. 
Furthermore, the performance of traditional non-MFR models show that a separate solution for masked faces is needed to achieve competitive results on masked images face recognition.


\begin{table*}[]
\begin{center}
\caption{Ablation Study on Face Recognition Benchmarks in three different scenarios, Accuracy in \%: The proposed MaskInv models are capable of performing well on unmasked data and suffer only a small drop in accuracy. The MaskInv-HG and MaskInv-LG are the top two performers on most benchmarks for both the masked vs non-masked and masked vs masked scenarios.}
\label{abl-bench}
\begin{tabular}{|l|c|c|c|c|c|}
\hline
\multicolumn{1}{|c|}{\textbf{No Masks}}             & LFW   & CFP-FP & AgeDB-30 & CALFW & CPLFW \\ \hline
ArcFace \cite{arcface}                              & 99.82 & 98.27     & 98.15      & 95.45 & 92.08 \\ \hline
MagFace \cite{magface}                              & \textbf{99.83} & 98.46  & 98.17    & 96.15 & 92.87 \\ \hline
ElasticFace-Arc  (teacher) \cite{elasticface}       & 99.80 & \textbf{98.67}  & \textbf{98.35}    & \textbf{96.17} & \textbf{93.27} \\ \hline
ElasticFace-Arc-Aug (ours)                          & 99.80 & 97.61  & 97.68    & 96.08 & 92.40 \\ \hline
MaskInv-LG (ours)                                   & 99.82 & 97.53  & 97.83    & 96.05 & 92.65 \\ \hline
MaskInv-HG (ours)                                   & 99.82 & 97.57  & 97.83    & 96.10 & 92.82 \\ \hline
\multicolumn{1}{|c|}{\textbf{Masked vs Non-Masked}}       &       &        &          &       &       \\ \hline
ArcFace \cite{arcface}                              &  99.45    &  95.50      & 94.68         &    94.76   &  89.83 \\ \hline
MagFace \cite{magface}                              &   99.33    & 94.00       &  94.12       &  94.15    & 88.67 \\ \hline
ElasticFace-Arc (teacher) \cite{elasticface}        & 99.40             & 95.29             & 95.38             & 94.42             & 90.40 \\ \hline
ElasticFace-Arc-Aug (ours)                          & 99.58             & 95.91             & 96.73             & 95.58             & 91.08 \\ \hline
MaskInv-LG (ours)                                   & 99.67             & 96.83             & \textbf{97.07}    & \textbf{95.90}    & \textbf{91.98} \\ \hline
MaskInv-HG (ours)                                   & \textbf{99.73}    & \textbf{96.94}    & 96.69             & 95.75             & 91.67 \\ \hline
\multicolumn{1}{|c|}{\textbf{Masked vs Masked}}                &       &        &          &       &       \\ \hline
ArcFace \cite{arcface}                              & 99.00             &  92.01            & 92.37             &  93.32            & 87.07 \\ \hline
MagFace \cite{magface}                              & 98.47             & 89.26             & 91.82             & 92.62             & 84.72  \\ \hline
ElasticFace-Arc (teacher) \cite{elasticface}        & 98.90             & 92.01             & 93.47             & 93.73             & 87.55 \\ \hline
ElasticFace-Arc-Aug (ours)                          & 99.55             & 96.07             & 95.08             & 95.25             & 90.10 \\ \hline
MaskInv-LG (ours)                                   & \textbf{99.57}    & \textbf{96.29}    & \textbf{96.51}    & \textbf{95.45}    & \textbf{91.45} \\ \hline
MaskInv-HG (ours)                                   & \textbf{99.57}    & \textbf{96.29}    & 96.47             & 95.42             & 91.32 \\ \hline
\end{tabular}
\end{center}
\vspace{-7mm}
\end{table*}

\vspace{-2mm}
\subsection{Comparison with SOTA}
\begin{table}[t]
\begin{center}
\caption{Comparison with SOTA on the MFRC-21 Dataset -  Mask vs No-Mask. Our MaskInv-HG model outperforms the top-performing academic solutions on the FMR1000 and the FDR metrics. On the FMR100, our model is placed second.}
\label{sota-mfrc21nomask}
\begin{tabular}{l|c|c|c|}
\cline{1-4}

\multicolumn{1}{|c|}{Solution}          & FMR1000          & FMR100           & FDR              \\ \hline
\multicolumn{1}{|l|}{MaskInv-HG (ours)} & \textbf{0.05849} & 0.05637          & \textbf{11.3271} \\ \hline \hline
\multicolumn{1}{|l|}{MTArcFace \cite{BoutrosMFR21}}         & 0.05860          & 0.05860          & 10.7497          \\ \hline
\multicolumn{1}{|l|}{MaskedArcFace \cite{BoutrosMFR21}}     & 0.05963          & 0.05687          & 10.4484          \\ \hline
\multicolumn{1}{|l|}{SMT-MFR-1 \cite{BoutrosMFR21}}         & 0.06003          & 0.05704          & 10.6824          \\ \hline
\multicolumn{1}{|l|}{LMI-SMT-MFR-1 \cite{BoutrosMFR21}}     & 0.06205          & 0.05722          & 9.7384           \\ \hline
\multicolumn{1}{|l|}{SMT-MFR-2 \cite{BoutrosMFR21}}         & 0.06268          & \textbf{0.05584} & 11.2025          \\ \hline
\multicolumn{1}{|l|}{VIPLFACE-M \cite{BoutrosMFR21}}        & 0.06279          & 0.05681          & 8.2371           \\ \hline
\multicolumn{1}{|l|}{LMI-SMT-MFR-2 \cite{BoutrosMFR21}}     & 0.07096          & 0.05848          & 8.5278           \\ \hline
\multicolumn{1}{|l|}{VIPLFACE-G \cite{BoutrosMFR21}}        & 0.07269          & 0.05750          & 8.1693           \\ \hline
\multicolumn{1}{|l|}{MFR-NMRE-B \cite{BoutrosMFR21}}        & 0.08344          & 0.05819          & 7.9504           \\ \hline
\multicolumn{1}{|l|}{MFR-NMRE-F \cite{BoutrosMFR21}}        & 0.17660          & 0.08125          & 5.3876           \\ \hline
\multicolumn{1}{|l|}{Neto et al. \cite{pedro}}             & -                & 0.28252          & -          \\ \hline
\end{tabular}
\end{center}
\vspace{-7mm}
\end{table}

\begin{table}[t]
\begin{center}
\caption{Comparison with SOTA on the MFRC-21 Dataset -  Mask vs Mask. Our MaskInv-HG model outperforms all top-performing academic solution on all evaluation metrics.}
\label{sota-mfrc21mask}
\begin{tabular}{l|c|c|c|}
\cline{1-4}
\multicolumn{1}{|c|}{Solution}          & FMR1000          & FMR100           & FDR              \\ \hline
\multicolumn{1}{|l|}{MaskInv-HG (ours)} & \textbf{0.05886} & \textbf{0.05654} & \textbf{11.5337} \\ \hline \hline
\multicolumn{1}{|l|}{SMT-MFR-1 \cite{BoutrosMFR21}}         & 0.06012          & 0.05825          & 11.0344          \\ \hline
\multicolumn{1}{|l|}{LMI-SMT-MFR-1 \cite{BoutrosMFR21}}     & 0.06061          & 0.05856          & 9.9091           \\ \hline
\multicolumn{1}{|l|}{SMT-MFR-2 \cite{BoutrosMFR21}}         & 0.06172          & 0.05792          & 11.3090          \\ \hline
\multicolumn{1}{|l|}{MaskedArcFace \cite{BoutrosMFR21}}     & 0.06245          & 0.05825          & 10.5731          \\ \hline
\multicolumn{1}{|l|}{VIPLFACE-G \cite{BoutrosMFR21}}        & 0.06359          & 0.05843          & 9.4147           \\ \hline
\multicolumn{1}{|l|}{MTArcface \cite{BoutrosMFR21}}         & 0.06390          & 0.05850          & 10.1700          \\ \hline
\multicolumn{1}{|l|}{LMI-SMT-MFR-2 \cite{BoutrosMFR21}}     & 0.06586          & 0.05916          & 8.8742           \\ \hline
\multicolumn{1}{|l|}{VIPLFACE-M \cite{BoutrosMFR21}}        & 0.06788          & 0.05759          & 8.9859           \\ \hline
\multicolumn{1}{|l|}{MFR-NMRE-B \cite{BoutrosMFR21}}        & 0.12903          & 0.05970          & 8.1196           \\ \hline
\multicolumn{1}{|l|}{MFR-NMRE-F \cite{BoutrosMFR21}}        & 0.19890          & 0.09630          & 4.7322           \\ \hline
\multicolumn{1}{|l|}{Neto et al. \cite{pedro}}             & -                & 0.23507          & -                \\ \hline
\end{tabular}
\end{center}
\vspace{-4mm}
\end{table}

In this subsection, we compare our proposed approach with previously published results on both the MFRC-21 \cite{BoutrosMFR21} and MFR2 \cite{masktheface} datasets. Since that the ablation study proved the benefit  of our proposed MaskInv-HG solution, especially in terms of generalization, we compare the results of this solution to SOTA results. The comparison to SOTA is limited to these two benchmarks (out of 7 used) as the rest of the benchmarks do not have comparable results in the literature. For the MFRC-21 dataset, we limit ourselves for the evaluation to verification performance and, unlike the challenge organizers, do not consider model compactness because we do not propose a novel architecture in this work. In addition, we consider the top-ranked academic submissions to the challenge, as they provide a detailed description of their proposed approaches. We compare our solution to the ten top-ranked academic approaches in the MFRC-21 challenge \cite{BoutrosMFR21} on the masked vs non-masked (\autoref{sota-mfrc21nomask}) and the masked vs masked scenario (\autoref{sota-mfrc21mask}). A more detailed description of the different solutions in the MFRC-21 competition can be found in the competition paper \cite{BoutrosMFR21}.

In both scenarios, our proposed MaskInv-HG outperformed the academic solutions submitted to the MFRC-21 challenges. In the masked vs non-masked (\autoref{sota-mfrc21nomask}) scenario, our model achieved a verification performance FMR1000 of 0.05849 and beats the best-performing academic solution MTArcFace that achieved a verification performance FMR1000 of 0.05860. As mention earlier, we focus our evaluation on the FMR1000 as it has been proposed as a best practice operation point for automatic border control by Frontex \cite{frontex2015best}. On the FMR100, only the SMT-MFR-2 model achieved a higher verification performance than our proposed MaskInv-HG model. In addition, our model achieves the highest FDR, which indicates a better separability of imposter and genuine distribution than the other approaches, and thus indicates higher generalizability. In the masked vs masked scenario (\autoref{sota-mfrc21mask}), our MaskInv-HG model outperforms the solutions proposed to the challenge on all evaluation metrics. In contrast to the other top-3 solutions on the masked vs non-masked scenario, the performance of our model on masked vs masked was only slightly worse (FMR1000 = 0.05886) than on the mask vs unmasked dataset (FMR1000 = 0.05849), regarding the FMR1000. This shows the flexibility of our proposed MaskInv-HG model to handle both scenarios well by creating similar face representations for masked and non-masked faces.

\begin{table}[t]
\begin{center}
\caption{Comparison with SOTA on the MFR2 Dataset: Our proposed MaskInv-HG model outperforms the model proposed by the authors of the dataset in each evaluation metric.}
\label{sota-mfr2}
\begin{tabular}{|l|c|c|c|}
\hline
\multicolumn{1}{|c|}{Solution}                    & FAR2000 & ACC2000 & ACC     \\ \hline 
MaskInv-HG (ours)                                 & \textbf{91.21\%} & \textbf{96.11\%} & \textbf{96.34\%} \\ \hline \hline
MTF Retrained \cite{masktheface} & 82.78\% & 91.18\% & 95.99\% \\ \hline
\end{tabular}
\end{center}
\vspace{-11mm}
\end{table}

In \autoref{sota-mfr2}, we compare our results on the MFR2 dataset with the proposed solution "MTF Retrained" of the authors of the dataset \cite{masktheface}. The "MTF Retrained" is based on the Inception-ResNet v1 \cite{inception} and is trained on a subset from the VGGFace2 dataset \cite{vggface2} with the triplet-loss FaceNet \cite{facenet}. On the images of this subset, different synthetic masks were added. In all three evaluation metrics, our MaskInv-HG model outperformed the MTF-Retrained model. Our model increased the performance by $8.43$ percentage points of the FAR2000, $4.93$ percentage points of the corresponding accuracy at FAR2000 (ACC2000), and the maximum accuracy from $95.99\%$ to $96.34\%$.


In summary, our proposed approach outperformed the previous SOTA in both the MFRC-21 challenge and the MFR2 evaluation protocol. This proves the ability of the proposed MaskInv solution to adapt to masked faces and provide a step forward towards accurate masked face recognition in both scenarios, masked vs masked and masked vs non-masked faces.

\vspace{-5.5mm}
\section{Conclusion}
\label{concl}
\vspace{-1mm}
In this paper, we proposed a mask-invariant face recognition solution that does not only aim at building discriminant face embeddings, but rather extend this target to building embeddings that maintain their intra-identity similarity whether wearing or not wearing a mask.
This novel approach, namely the MaskInv, is based on jointly learning to separate between identities despite wearing masks through identity classification learning, and learning to produce similar embeddings for masked and non-masked faces of the same identity through embedded level KD from a teacher network. 
In a detailed ablation study, on two real masked datasets as well as on five mainstream face verification benchmarks, different stages of our proposed approach have shown to enhance the performance of masked face recognition. 
The proposed solution outperformed current SOTA approaches when compared to the academic solutions submitted to the recent masked face recognition competition MFRC-21. 
Moreover, our proposed solution maintains high-level of accuracy when verifying non-masked faces as we demonstrated show on five widely used benchmarks addressing different variations including cross-pose and cross-age verification. As this and future pandemic foreseen effect on the world is still unknown, masked face recognition, which has been brought into focus by the global COVID-19 pandemic, will be of further interest for our society and require novel solutions. Additionally, the potential of the presented mask-invariant face representation learning may not be limited to its use for masked face recognition, but could also prove useful for the problem of occluded face recognition in general, which is still to be studied.

\vspace{-5mm}
{\small
\bibliographystyle{ieee}
\bibliography{egbib}

\begin{thebibliography}{10}\itemsep=-1pt

\bibitem{masktheface}
A.~Anwar and A.~Raychowdhury.
\newblock Masked face recognition for secure authentication.
\newblock {\em CoRR}, abs/2008.11104, 2020.

\bibitem{batagelj}
B.~Batagelj, P.~Peer, V.~Štruc, and S.~Dobrišek.
\newblock How to correctly detect face-masks for covid-19 from visual
  information?
\newblock {\em Applied Sciences}, 11(5), 2021.

\bibitem{elasticface}
F.~Boutros, N.~Damer, F.~Kirchbuchner, and A.~Kuijper.
\newblock Elasticface: Elastic margin loss for deep face recognition.
\newblock {\em CoRR}, abs/2109.09416, 2021.

\bibitem{triplet}
F.~Boutros, N.~Damer, F.~Kirchbuchner, and A.~Kuijper.
\newblock Self-restrained triplet loss for accurate masked face recognition,
  2021.

\bibitem{BoutrosMFR21}
F.~Boutros, N.~Damer, J.~N. Kolf, K.~Raja, F.~Kirchbuchner, R.~Ramachandra,
  A.~Kuijper, P.~Fang, C.~Zhang, F.~Wang, D.~Montero, N.~Aginako, B.~Sierra,
  M.~Nieto, M.~E. Erakin, U.~Demir, H.~K. Ekenel, A.~Kataoka, K.~Ichikawa,
  S.~Kubo, J.~Zhang, M.~He, D.~Han, S.~Shan, K.~Grm, V.~Struc, S.~Seneviratne,
  N.~Kasthuriarachchi, S.~Rasnayaka, P.~C. Neto, A.~F. Sequeira, J.~R. Pinto,
  M.~Saffari, and J.~S. Cardoso.
\newblock {MFR} 2021: Masked face recognition competition.
\newblock In {\em {IJCB}}, pages 1--10. {IEEE}, 2021.

\bibitem{pocketnet}
F.~Boutros, P.~Siebke, M.~Klemt, N.~Damer, F.~Kirchbuchner, and A.~Kuijper.
\newblock Pocketnet: Extreme lightweight face recognition network using neural
  architecture search and multi-step knowledge distillation.
\newblock {\em CoRR}, abs/2108.10710, 2021.

\bibitem{vggface2}
Q.~Cao, L.~Shen, W.~Xie, O.~M. Parkhi, and A.~Zisserman.
\newblock Vggface2: {A} dataset for recognising faces across pose and age.
\newblock In {\em {FG}}, pages 67--74. {IEEE} Computer Society, 2018.

\bibitem{humans}
N.~Damer, F.~Boutros, M.~S{\"{u}}{\ss}milch, M.~Fang, F.~Kirchbuchner, and
  A.~Kuijper.
\newblock Masked face recognition: Human vs. machine.
\newblock {\em CoRR}, abs/2103.01924, 2021.

\bibitem{damer2021extended}
N.~Damer, F.~Boutros, M.~Süßmilch, F.~Kirchbuchner, and A.~Kuijper.
\newblock Extended evaluation of the effect of real and simulated masks on face
  recognition performance.
\newblock {\em IET Biometrics}, 10(5):548--561, 2021.

\bibitem{DamerGCBKK20}
N.~Damer, J.~H. Grebe, C.~Chen, F.~Boutros, F.~Kirchbuchner, and A.~Kuijper.
\newblock The effect of wearing a mask on face recognition performance: an
  exploratory study.
\newblock In {\em {BIOSIG}}, volume {P-306} of {\em {LNI}}, pages 1--10.
  Gesellschaft f{\"{u}}r Informatik e.V., 2020.

\bibitem{deng2021masked}
J.~Deng, J.~Guo, X.~An, Z.~Zhu, and S.~Zafeiriou.
\newblock Masked face recognition challenge: The insightface track report.
\newblock {\em CoRR}, abs/2108.08191, 2021.

\bibitem{arcface}
J.~Deng, J.~Guo, N.~Xue, and S.~Zafeiriou.
\newblock Arcface: Additive angular margin loss for deep face recognition.
\newblock In {\em {CVPR}}, pages 4690--4699. Computer Vision Foundation /
  {IEEE}, 2019.

\bibitem{FANG2021108398}
M.~Fang, N.~Damer, F.~Kirchbuchner, and A.~Kuijper.
\newblock Real masks and spoof faces: On the masked face presentation attack
  detection.
\newblock {\em Pattern Recognition}, page 108398, 2021.

\bibitem{frontex2015best}
Frontex.
\newblock Best practice technical guidelines for automated border control (abc)
  systems, 2015.

\bibitem{FuMaskedQuality}
B.~Fu, F.~Kirchbuchner, and N.~Damer.
\newblock The effect of wearing a face mask on face image quality.
\newblock {\em CoRR}, abs/2110.11283, 2021.

\bibitem{Ge2017CVPR}
S.~Ge, J.~Li, Q.~Ye, and Z.~Luo.
\newblock Detecting masked faces in the wild with lle-cnns.
\newblock In {\em {CVPR}}, pages 426--434. {IEEE} Computer Society, 2017.

\bibitem{kdlow}
S.~Ge, S.~Zhao, C.~Li, and J.~Li.
\newblock Low-resolution face recognition in the wild via selective knowledge
  distillation.
\newblock {\em IEEE Transactions on Image Processing}, 28(4), 2019.

\bibitem{gan2}
M.~Geng, P.~Peng, Y.~Huang, and Y.~Tian.
\newblock Masked face recognition with generative data augmentation and domain
  constrained ranking.
\newblock In {\em {ACM} Multimedia}, pages 2246--2254. {ACM}, 2020.

\bibitem{martaERA}
M.~Gomez{-}Barrero, P.~Drozdowski, C.~Rathgeb, J.~Patino, M.~Todisco,
  A.~Nautsch, N.~Damer, J.~Priesnitz, N.~W.~D. Evans, and C.~Busch.
\newblock Biometrics in the era of {COVID-19:} challenges and opportunities.
\newblock {\em CoRR}, abs/2102.09258, 2021.

\bibitem{ms1m}
Y.~Guo, L.~Zhang, Y.~Hu, X.~He, and J.~Gao.
\newblock Ms-celeb-1m: {A} dataset and benchmark for large-scale face
  recognition.
\newblock In {\em {ECCV} {(3)}}, volume 9907 of {\em Lecture Notes in Computer
  Science}. Springer, 2016.

\bibitem{resnet100}
K.~He, X.~Zhang, S.~Ren, and J.~Sun.
\newblock Deep residual learning for image recognition.
\newblock In {\em {CVPR}}, pages 770--778. {IEEE} Computer Society, 2016.

\bibitem{distilling}
G.~E. Hinton, O.~Vinyals, and J.~Dean.
\newblock Distilling the knowledge in a neural network.
\newblock {\em CoRR}, abs/1503.02531, 2015.

\bibitem{lfw}
G.~B. Huang, M.~Ramesh, T.~Berg, and E.~Learned-Miller.
\newblock Labeled faces in the wild: A database for studying face recognition
  in unconstrained environments.
\newblock Technical Report 07-49, University of Massachusetts, Amherst, October
  2007.

\bibitem{curricularface}
Y.~Huang, Y.~Wang, Y.~Tai, X.~Liu, P.~Shen, S.~Li, J.~Li, and F.~Huang.
\newblock Curricularface: Adaptive curriculum learning loss for deep face
  recognition.
\newblock In {\em {CVPR}}, pages 5900--5909. Computer Vision Foundation /
  {IEEE}, 2020.

\bibitem{JainRP04}
A.~K. Jain, A.~Ross, and S.~Prabhakar.
\newblock An introduction to biometric recognition.
\newblock {\em {IEEE} Trans. Circuits Syst. Video Technol.}, 14(1):4--20, 2004.

\bibitem{JiaM09}
H.~Jia and A.~M. Mart{\'{\i}}nez.
\newblock Support vector machines in face recognition with occlusions.
\newblock In {\em {CVPR}}, pages 136--141. {IEEE} Computer Society, 2009.

\bibitem{LiMasks}
C.~Li, S.~Ge, D.~Zhang, and J.~Li.
\newblock Look through masks: Towards masked face recognition with de-occlusion
  distillation.
\newblock In {\em {ACM} Multimedia}, pages 3016--3024. {ACM}, 2020.

\bibitem{li_attention}
Y.~Li, K.~Guo, Y.~Lu, and L.~Liu.
\newblock Cropping and attention based approach for masked face recognition.
\newblock {\em Appl. Intell.}, 51(5):3012--3025, 2021.

\bibitem{loey}
M.~Loey, G.~Manogaran, M.~H.~N. Taha, and N.~E.~M. Khalifa.
\newblock A hybrid deep transfer learning model with machine learning methods
  for face mask detection in the era of the covid-19 pandemic.
\newblock {\em Measurement}, 167, 2021.

\bibitem{mansfield2006information}
A.~Mansfield.
\newblock Information technology--biometric performance testing and
  reporting--part 1: Principles and framework.
\newblock {\em ISO/IEC}, 2006.

\bibitem{magface}
Q.~Meng, S.~Zhao, Z.~Huang, and F.~Zhou.
\newblock Magface: {A} universal representation for face recognition and
  quality assessment.
\newblock In {\em {CVPR}}, pages 14225--14234. Computer Vision Foundation /
  {IEEE}, 2021.

\bibitem{agedb}
S.~Moschoglou, A.~Papaioannou, C.~Sagonas, J.~Deng, I.~Kotsia, and
  S.~Zafeiriou.
\newblock Agedb: The first manually collected, in-the-wild age database.
\newblock In {\em {CVPR} Workshops}, pages 1997--2005. {IEEE} Computer Society,
  2017.

\bibitem{margin}
D.~Nekhaev, S.~Milyaev, and I.~Laptev.
\newblock Margin based knowledge distillation for mobile face recognition.
\newblock In {\em {ICMV}}, {SPIE} Proceedings. {SPIE}, 2019.

\bibitem{pedro}
P.~C. Neto, F.~Boutros, J.~R. Pinto, M.~Saffari, N.~Damer, A.~F. Sequeira, and
  J.~S. Cardoso.
\newblock My eyes are up here: Promoting focus on uncovered regions in masked
  face recognition.
\newblock {\em BIOSIG}, abs/2108.00996, 2021.

\bibitem{FRVT}
M.~L. Ngan, P.~J. Grother, and K.~K. Hanaoka.
\newblock Ongoing face recognition vendor test (frvt) part 6b: Face recognition
  accuracy with face masks using post-covid-19 algorithms.
\newblock NIST Interagency/Internal Report (NISTIR) - 8331, 2020.

\bibitem{fdr}
N.~Poh and S.~Bengio.
\newblock A study of the effects of score normalisation prior to fusion in
  biometric authentication tasks.
\newblock Technical report, IDIAP, 2004.

\bibitem{facenet}
F.~Schroff, D.~Kalenichenko, and J.~Philbin.
\newblock Facenet: {A} unified embedding for face recognition and clustering.
\newblock In {\em {CVPR}}, pages 815--823. {IEEE} Computer Society, 2015.

\bibitem{cfp}
S.~Sengupta, J.~Chen, C.~D. Castillo, V.~M. Patel, R.~Chellappa, and D.~W.
  Jacobs.
\newblock Frontal to profile face verification in the wild.
\newblock In {\em {WACV}}, pages 1--9. {IEEE} Computer Society, 2016.

\bibitem{inception}
C.~Szegedy, S.~Ioffe, V.~Vanhoucke, and A.~A. Alemi.
\newblock Inception-v4, inception-resnet and the impact of residual connections
  on learning.
\newblock In {\em {AAAI}}, pages 4278--4284. {AAAI} Press, 2017.

\bibitem{gan1}
N.~Ud~Din, K.~Javed, S.~Bae, and J.~Yi.
\newblock A novel gan-based network for unmasking of masked face.
\newblock {\em IEEE Access}, 8, 2020.

\bibitem{Wang_2019_ICCV}
M.~Wang, R.~Liu, H.~Nada, N.~Abe, H.~Uchida, and T.~Matsunami.
\newblock Improved knowledge distillation for training fast low resolution face
  recognition model.
\newblock In {\em {ICCV} Workshops}, pages 2655--2661. {IEEE}, 2019.

\bibitem{occ18}
C.~Wu and J.~Ding.
\newblock Occluded face recognition using low-rank regression with generalized
  gradient direction.
\newblock {\em Pattern Recognit.}, 80:256--268, 2018.

\bibitem{mtcnn}
K.~Zhang, Z.~Zhang, Z.~Li, and Y.~Qiao.
\newblock Joint face detection and alignment using multitask cascaded
  convolutional networks.
\newblock {\em IEEE Signal Processing Letters}, 23(10), 2016.

\bibitem{cplfw}
T.~Zheng and W.~Deng.
\newblock Cross-pose lfw: A database for studying cross-pose face recognition
  in unconstrained environments.
\newblock Technical Report 18-01, Beijing University of Posts and
  Telecommunications, February 2018.

\bibitem{calfw}
T.~Zheng, W.~Deng, and J.~Hu.
\newblock Cross-age {LFW:} {A} database for studying cross-age face recognition
  in unconstrained environments.
\newblock {\em CoRR}, abs/1708.08197, 2017.

\bibitem{web}
W.~Zheng, L.~Yan, F.~Wang, and C.~Gou.
\newblock Learning from the web: Webly supervised meta-learning for masked face
  recognition.
\newblock In {\em {CVPR} Workshops}, pages 4304--4313. Computer Vision
  Foundation / {IEEE}, 2021.

\bibitem{zhu2021masked}
Z.~Zhu, G.~Huang, J.~Deng, Y.~Ye, J.~Huang, X.~Chen, J.~Zhu, T.~Yang, J.~Guo,
  J.~Lu, D.~Du, and J.~Zhou.
\newblock Masked face recognition challenge: The webface260m track report.
\newblock {\em CoRR}, abs/2108.07189, 2021.

\bibitem{vitomir}
V.~Štruc, S.~Dobrišek, and N.~Pavešić.
\newblock Confidence weighted subspace projection techniques for robust face
  recognition in the presence of partial occlusions.
\newblock In {\em 2010 20th ICPR}, 2010.

\end{thebibliography}
}

\end{document}